\documentclass[final,5p,times,twocolumn]{elsarticle}

\usepackage{hyperref} % 超链接包
\usepackage{amsmath}
\usepackage{multirow}  %multirow for format of table 
\usepackage{xcolor}
\usepackage{float}
\usepackage{booktabs} % 表格包
\usepackage{graphicx}  % 图片包
\usepackage{subfigure}  % 子图包
\usepackage{algorithm} % format of the algorithm 
\usepackage{algorithmic} % format of the algorithm 

\def\var{\mbox{Var}}

\newtheorem{definition}{Definition} 
\newcommand{\myeqref}[1]{Eq. \ref{#1}}
\newcommand{\myfigref}[1]{Fig. \ref{#1}}

\newcommand{\myalgref}[1]{Alg. \ref{#1}}
\newcommand{\mytabref}[1]{Table \ref{#1}}
\hypersetup{
	colorlinks=true,
	linkcolor=blue,
	filecolor=magenta,      
	urlcolor=cyan,
	pdftitle={Overleaf Example},
	pdfpagemode=FullScreen,
}

\journal{Information Sciences}

\begin{document}

\begin{frontmatter}

\title{An Improved Probability Propagation Algorithm for Density Peak Clustering Based on Natural Nearest Neighborhood}

\author[1]{Wendi Zuo}
\ead{zuowendi@mail.ustc.edu.cn}

\author[1,2,3]{Xinmin Hou\corref{cor1}}
\ead{xmhou@ustc.edu.cn}

\cortext[cor1]{Corresponding author}

\address[1]{School of Big Data, University of Science and Technology of China, Hefei, Anhui 230026, China}
\address[2]{School of Mathematical Sciences, University of Science and Technology of China, Hefei, Anhui 230026, China}
\address[3]{CAS Key Laboratory of Wu Wen-Tsun Mathematics, University of Science and Technology of China, Hefei, Anhui 230026, China}

\begin{abstract}
Clustering by fast search and find of density peaks (DPC) (Since, 2014) has been proven to be a promising clustering approach that efficiently discovers the centers of clusters by finding the density peaks. The accuracy of DPC depends on the cutoff distance ($d_c$), the cluster number ($k$) and the selection of the centers of clusters. Moreover, the final allocation strategy is sensitive and has poor fault tolerance. The shortcomings above make the algorithm sensitive to parameters and only applicable for some specific datasets. To overcome the limitations of DPC, this paper presents an improved probability propagation algorithm for density peak clustering based on the natural nearest neighborhood (DPC-PPNNN). By introducing the idea of natural nearest neighborhood and probability propagation, DPC-PPNNN realizes the nonparametric clustering process and makes the algorithm applicable for more complex datasets. In experiments on several datasets, DPC-PPNNN is shown to outperform DPC, K-means and DBSCAN.	
\end{abstract}


\end{frontmatter}

\section{Introduction}
Clustering, also known as unsupervised classification, aims to divide datasets into subsets or clusters according to the similarity measure of the data sample (physical or abstract) such that the data samples within the subset or cluster have a high degree of similarity and that the data samples belonging to different subsets or clusters have a high degree of dissimilarity \cite{TSK2011}. Currently, cluster analysis plays an important role in many fields such as social sciences, biology, pattern recognition, information retrieval and so on \cite{LWY2018}. It is so useful in machine learning and data mining that many researchers have paid much attention to it. Over the past few decades, a number of excellent clustering algorithms have been developed for different types of applications. Typical algorithms include K-means \cite{MJ1967} and K-medoids \cite{KR1987} based on partitioning, CURE \cite{GRS1998} and BIRCH \cite{TRL1996} based on hierarchy, DBSCAN \cite{EKHSX1996} and OPTICS \cite{ABKS1999} based on density, WaveCluster \cite{SGCSZ2000} and STING \cite{WYR1997} based on grids and statistical clustering \cite{D1977} based on models.

In 2014, a density-based clustering algorithm DPC was given by Rodriguez and Laio \cite{RL2014} (Clustering by fast search and find of density peaks, Science, 344~(2014) 1492). The main idea of the algorithm DPC is as follows: For each data point $i$, we compute two quantities: its local density $\rho_i$ and its distance $\delta_i$ from points of higher density. By mapping all the data points to the decision graph which takes $\rho$ and $\delta$ as the two axes, we can recognize density peak points (cluster centers) as points for which the values of $\rho_i$ and $\delta_i$ are anomalously large. After the cluster centers have been found, each remaining point is assigned to the same cluster as its nearest neighbor of higher density. The DPC algorithm is simple and efficient, and it can quickly find the high density peak points (cluster centers) without iteratively calculating the objective function. Moreover, it is suitable for cluster analysis on large-scale data. Although the DPC algorithm has obvious advantages over other clustering algorithms, it also has some shortcomings: the accuracy of DPC depends on the cutoff distance ($d_c$), the cluster number ($k$) and the selection of the centers of clusters. Moreover, the final allocation strategy is sensitive and has poor fault tolerance. Last, the algorithm has its basis on the assumptions that cluster centers are surrounded by neighbors with lower local density and that they are at a relatively large distance from any point with higher local density.

To avoid the deficiencies of DPC, in this paper, we present an improved probability propagation algorithm for density peak clustering based on natural nearest neighborhood (DPC-PPNNN). By introducing the concept of natural nearest neighborhood, we can avoid setting $d_c$ manually. By calculating $\gamma=\rho\times\delta$ and $\theta=\frac{\rho}{\delta}$, we can select cluster centers automatically. Finally, the clustering algorithm based on probability propagation can help us allocate all the remaining data points. By doing all these, DPC-PPNNN is suitable for more complex datasets and can distinguish two clusters that are close to each other.

The rest of this paper is organized as follows. In Section \ref{related-works}, we introduce the research progress related to the DPC algorithm. In Section \ref{DPC and analysis}, we briefly review the basic definitions and processes of the traditional DPC algorithm and reveal some problems within them. In Section \ref{NNNsection}, we introduce the basic concept and the algorithm of the natural nearest neighborhood. In Section \ref{selection-of-cluster-centers}, we show the improvement on the local density and the method of selecting cluster centers automatically. In Section \ref{PPsection}, we propose the allocation algorithm based on probability propagation. In Section \ref{experiment-section}, we compare the DPC-PPNNN algorithm with other classical clustering algorithms using various synthetic and real-world datasets. In Section \ref{conclusions-section}, we summarize the advantages and disadvantages of the DPC-PPNNN algorithm and point out the direction of our future research.

\section{Related works} \label{related-works}
DPC has been proven to be an effective clustering strategy but it also has many limitations such as the arbitrary selection of density estimation metrics, clustering centers and the risk of error propagation. Over the past few years, many optimized variants of DPC have been proposed considering the following aspects:

The first aspect is improving the density measure of the DPC algorithm. Du, Ding and Jia~\cite{DDJ2016} proposed DPC-KNN, which introduces the concept of K-nearest neighbors (KNN) to DPC and provides another option for computing the local density. They also employ PCA to reduce the dimensionality of data. However, the method still suffers from the limitations of DPC because it applies the same procedure in determining the cluster centers and assigning the rest of the points. Liu, Wang, and Yu~\cite{LWY2018}  proposed SNN-DPC, which estimates the local density based on the idea of shared nearest neighbor similarity. However, it still suffers from the limitation of manually selecting the number of cluster centers. Mehmood et al~\cite{MZBDA2016} proposed CFSFDP-HD, which applies a nonparametric density estimator (Kernel Density Estimation, KDE) to eliminate the reliance of DPC on the cut-off distance $d_c$.

The second aspect is to automatically recognize the numbers of clusters and cluster centers. Liang and Chen~\cite{LZC2016} proposed 3DC, which introduces a divide-and-conquer strategy to determine the ideal number of clusters. However, it ignores the local structure of the datasets which may cause missing clusters. Xu, Wang and Deng~\cite{XWD2016} proposed DenPEHC, which could automatically detect all possible centers and build a hierarchy presentation for the dataset. Nevertheless, both 3DC and DenPEHC will aggravate the propagation of errors due to the hierarchical clustering strategy. Li, Ge, and Su~\cite{THS2016} proposed an automatic clustering algorithm for determining the density of clustering centers. In this algorithm, it is considered that if the shortest distance between a potential cluster center and a known cluster center is less than the cutoff distance $d_c$, then the potential cluster center is a redundant center. Otherwise, it is regarded as the actual center of another cluster.

The third aspect is to improve the assignment strategy to reduce error propagation. In most of the DPC variants, the idea of K-nearest neighbors is hybridized in the aggregation strategies. For instance, Zhou et al~\cite{ZSZZ2018} constructed the veins of clusters by connecting pairs with the highest similarity from the high-density regions to the cluster borders. The rest of the points are then assigned to the nearest veins. Xie et al~\cite{XGXLG2016} proposed a two-step procedure for label propagation. The first strategy assigns non-outliers based on the search of the K-nearest neighbors starting from the density peaks. The second strategy assigns the other points using the fuzzy KNN technique. Geng et al~\cite{GLZZH2018} proposed a KNN graph-based label propagation strategy to assign the remaining points. Liu, Wang and Yu~\cite{LWY2018} introduced a two-step allocation method based on inevitably and possibly subordinate allocation of the noncenter points.

\section{DPC algorithm and analysis} \label{DPC and analysis}
\subsection{DPC algorithm}
The DPC algorithm presented by Rodriguez and Laio~\cite{RL2014} has its basis on the assumption that cluster centers are surrounded by neighbors with lower local density and that they are at a relatively larger distance from any point with higher local density. There are two quantities to describe each data point $i$: its local density $\rho_i$ and its distance $\delta_i$ from points of higher density.

The DPC algorithm provides two methods for calculating the local density for a data point $i$: the cutoff distance method and the kernel distance method. For a data point $i$, its local density $\rho_i$ is defined as 
\begin{equation} \label{cutoff-density}
	\rho_i = \sum_{i \neq j} \chi (d_{ij}-d_c), \quad \chi (x) = 
	\begin{cases}
		1, \quad x < 0 \\
		0, \quad x > 0 \\
	\end{cases}
\end{equation}
with the cutoff distance method, or
\begin{equation} \label{kernel-density}
	\rho_i = \sum_{i \neq j} \exp \left[ - {\left( \frac{d_{ij}}{d_c} \right) } ^2 \right]
\end{equation}
with the kernel distance method, where $d_{ij}$ is the Euclidean distance between data points $i$ and $j$, $d_c(>0)$, the cutoff distance, is the radius of a point for scanning its neighborhood, which is set by the user. Thus, the local density $\rho_i$ is positively correlated with the number of points with a distance from $i$ less than $d_c$. The most obvious difference between the two methods is that for \myeqref{cutoff-density}, $\rho_i$ is a discrete value, whereas for  \myeqref{kernel-density}, it is a continuous value. However, for both methods, $\rho_i$ is sensitive to $d_c$.

Subsequently, $\delta_i$ in DPC is defined as 
\begin{equation} \label{distance1}
	\delta_i = \min \limits_{j:\rho_j > \rho_i} \left( d_{ij} \right)
\end{equation}
For the data point $i$ that has the highest local density $\rho_i$, $\delta_i$ is conventionally defined as
\begin{equation} \label{distance2}
	\delta_i = \max \limits_j \left( d_{ij} \right)
\end{equation}

As shown in \myeqref{distance1}, $\delta_i$ is the minimum distance between point $i$ and points $j$ with local densities $\rho_j>\rho_i$. 

After calculating the two quantities: $\rho$ and $\delta$, we can recognize density peak points (cluster centers) as points for which the values of $\rho_i$ and $\delta_i$ are anomalously large by mapping all the data points to the decision graph which takes $\rho$ and $\delta$ as the two axes. However, sometimes we cannot select cluster centers from the decision graph correctly because they are too close to each other. Instead, we can sort all the data points by $\gamma$ defined in \myeqref{gamma} and choose the largest $k$ (the cluster number) data points as the cluster centers.
\begin{equation} \label{gamma}
	\gamma_i = \rho_i \times \delta_i
\end{equation}

Since the cluster centers have been found, each remaining point is assigned to the same cluster as its nearest neighbor of higher density.

\subsection{Analysis} \label{analysis}
Although the experimental results obtained with DPC have shown that DPC performs well in many instances, the following drawbacks are obvious.

\begin{figure*}[htbp] % 插入图片dc例子
	\centering
	\subfigure[dc=0.0008]{
		\begin{minipage}[t]{0.3\linewidth}
			\centering
			\includegraphics[width=1.8in]{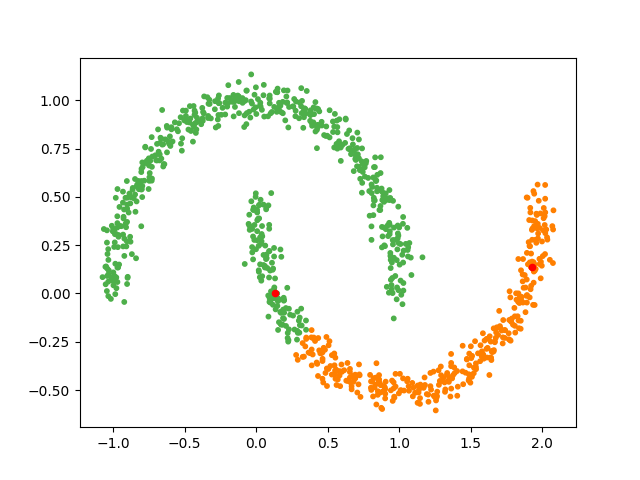}
	\end{minipage}}
	\subfigure[dc=0.0091]{
		\begin{minipage}[t]{0.3\linewidth}
			\centering
			\includegraphics[width=1.8in]{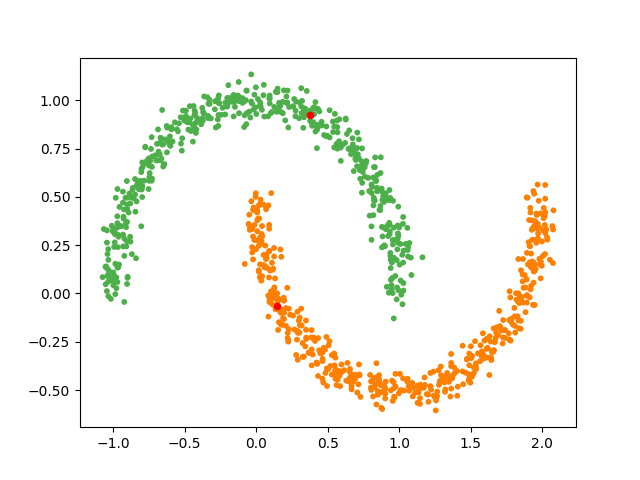}
	\end{minipage}}
	\subfigure[dc=0.4200]{
		\begin{minipage}[t]{0.3\linewidth}
			\centering
			\includegraphics[width=1.8in]{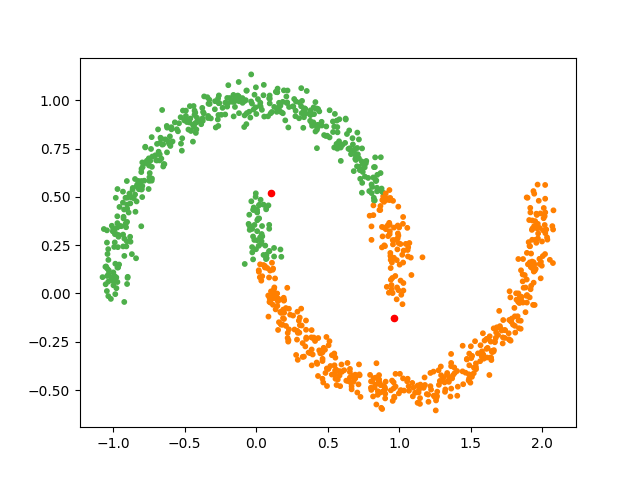}
	\end{minipage}}
	\caption{Results of the traditional DPC algorithm on the two-moon dataset}
	\label{dc_samples}
\end{figure*}

First, the accuracy of DPC depends on the setting of the cutoff distance ($d_c$). As shown in \myfigref{dc_samples}, there is a two-moon dataset. The red points are the cluster centers we select and the same for all the other pictures in this article. When we change the value of $d_c$, the clustering result of DPC can be greatly affected even for this simple dataset.

\begin{figure*}[htbp]  % 插入图片k例子
	\centering
	\subfigure[k=2]{
		\begin{minipage}[t]{0.4\linewidth}
			\centering
			\includegraphics[width=2in]{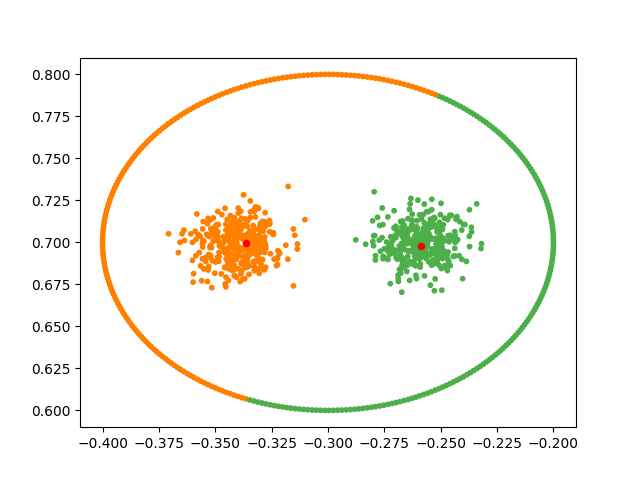}
			\label{k=2}
	\end{minipage}}
	\subfigure[k=3]{
		\begin{minipage}[t]{0.4\linewidth}
			\centering
			\includegraphics[width=2in]{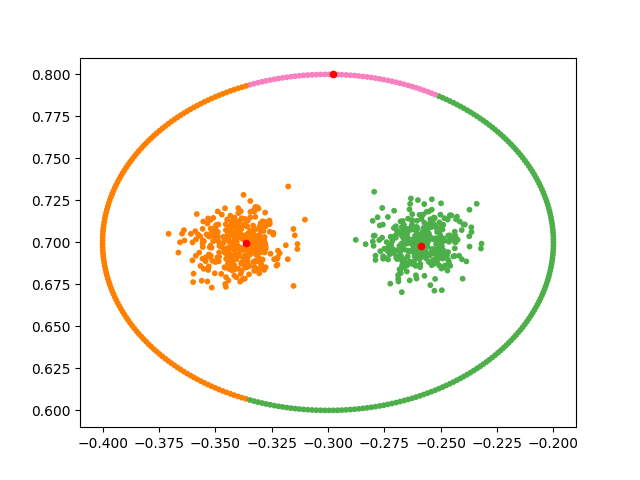}
			\label{k=3}
	\end{minipage}}
	\caption{Results of the traditional DPC algorithm on the Donut3 dataset}
	\label{k_samples}
\end{figure*}

Second, the cluster number ($k$) is difficult to determine. Actually, we  have little idea about the distribution of the dataset most of the time, not to mention choosing an ideal cluster number ($k$) for DPC. \myfigref{k_samples} shows the results of the traditional DPC algorithm on the Donut3 dataset. In \myfigref{k=2}, we recognize the outer ring as noise and let the cluster number be 2. However, clearly, the result cannot satisfy us. In \myfigref{k=3}, we recognize the outer ring as a cluster and let the cluster number be 3. Even when we choose the correct cluster center in the outer ring, we cannot obtain the ideal result.

\begin{figure*}[htbp] 
	\centering
	\includegraphics[width=3in]{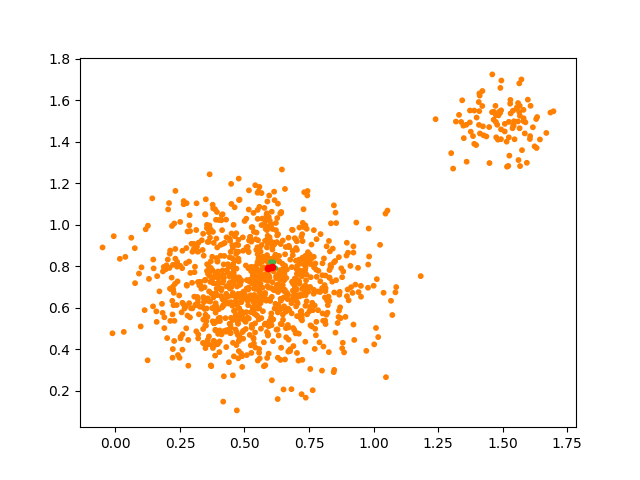}
	\caption{Results of the traditional DPC algorithm on the dataset with two Gaussian clusters with very different densities}
	\label{center_samples}
\end{figure*}

Third, it is difficult to recognize the centers of clusters. As shown in \myfigref{center_samples}, there is a dataset that has two clusters  based on a Gaussian distribution with one of the densities much larger than the other. Even when we choose the two data points with the largest $\gamma_i$ as the cluster centers, both of them will belong to the same cluster, which has a larger density. The reason for this phenomenon is that the difference in densities between the two clusters is so large that we can hardly recognize the centers of clusters regardless of no matter by the decision graph or by sorting all the data points by $\gamma$.
% 插入图片center例子

Last, the final allocation strategy is sensitive and has poor fault tolerance. As shown in \myfigref{k=3}, even when we choose the three cluster centers correctly, we cannot obtain the ideal result. There are still some data points that cannot be allocated to the correct cluster.

To overcome the deficiencies mentioned above, in this article, we present an improved probability propagation algorithm for density peak clustering based on natural nearest neighbors (DPC-PPNNN). By introducing the concept of natural nearest neighbors, we can avoid setting $d_c$ manually. By calculating $\gamma=\rho\times\delta$ and $\theta=\frac{\rho}{\delta}$, we can select cluster centers automatically. Finally, the clustering algorithm based on probability propagation can help us allocate all the remaining data points. By doing all these, DPC-PPNNN can be suitable for more complex datasets and distinguish two clusters that are close to each other.

\section{DPC-PPNNN}
\subsection{Natural Nearest Neighborhood} \label{NNNsection}
The natural nearest neighborhood (NNN) \cite{ZHFZ2014} is very different from traditional $k$-nearest neighbors. The NNN is a scale-free concept and does not require parameters in the selection of neighbors. The size of NNN of every data point may be different according to the distribution of the dataset. The following is a precise description of the natural neighborhood method through the definition of the relevant concepts.

Let $X=\{x_1,x_2,...,x_n\}\subset \mathbb{R}^d$ be a dataset. We use the Euclidean distance $d(x_i, x_j)$ to measure the distance between points $x_i$ and $x_j$. For a given integer $k>0$, let $nn_k(x_i)$ be the $k$-th nearest neighbor of $x_i$ in $X\setminus\{x_i\}$.  Define the {\em $k$-nearest neighborhood} of $x_i$ ($NN_k(x_i)$) as the set of $k$ nearest neighbors of $x_i$, i.e. 
$$NN_k(x_i)=\cup_{j=1}^k\{nn_j(x_i)\}.$$
The {\it $k$-reverse nearest neighborhood} of $x_i$ ($RNN_{k}(x_i)$)  is the set of points $x_j\in X\setminus\{x_i\}$ with $x_i\in NN_k(x_j)$, i.e. 
$$RNN_{k}(x_i) = \{ x_j \in X\setminus\{x_i\} \enspace | \enspace x_i \in NN_k(x_j) \}.$$  
The $k$-natural nearest neighborhood of $x_i$ ($NNN_k(x_i)$) is defined as 
$$NNN_k(x_i)=NN_k(x_i)\cap RNN_k(x_i).$$
Clearly, $NNN_{n-1}(x_i)=X\setminus\{x_i\}\not=\emptyset$ for any $x_i\in X$. 

\begin{definition}[Natural Nearest Neighborhood ($NNN$)]\label{NNN}
The least integer $\lambda$ with $NNN_\lambda(x_i)\not=\emptyset$ for every $x_i\in X$ is called the natural eigenvalue of $X$. We define $NNN_\lambda(x_i)$  as the natural nearest neighborhood of $x_i$,  denoted by $NNN(x_i)$, for every $x_i\in X$,. 
	% 	 The natural nearest neighbors searching process reaches natural stable state only if all points have at least one mutual neighbor, when the searching round $r$ increases from 1 to $\lambda$.
\end{definition}

Let $NNN_k(X)=\{NNN_k(x_i)\, |\, x_i\in X\}.$ Note that $NNN_k(X)$ maybe a multiset. Let $$NNN_k^0(X)=\{ x_j\, |\, x_j\in X \mbox{ with } NNN_k(x_j)=\emptyset\}.$$

By Definition~\ref{NNN}, the natural eigenvalue $\lambda$ represents the number of iterations in the process of searching the natural nearest neighborhood $NNN(X)=NNN_\lambda(X)$.
The natural eigenvalue $\lambda$ is generally small when there is no outlier or noise (the data points make $\lambda\in \Omega(\ln n)$). In the extremal case, $\lambda$ can be the largest possible value $n-1$  if there is an outlier $x_j\notin NN_{n-2}(x_i)$ for any $x_i\not=x_j$ and $x_i\in X$.

To decrease the effect of outliers or noise in dataset $X$, we introduce $\ln n + \ln \lambda$ as the threshold to control the number of iterations when $NNN_k^0(X)$ remains unchanged in its subsequent iteration, i.e. $NNN_k^0(X)=NNN_{k+1}^0(X)$. Formally, we define the logarithmic natural nearest neighborhood as follows:

\begin{definition}[Logarithmic Natural Nearest Neighborhood]\label{LNNN}
	We call the least integer $\lambda$ with 
	$$|\{i\, |\, i\le \lambda-1 \text{ with } NNN_i^0(X)=NNN_{i+1}^0(X) \}|\ge \ln n+\ln\lambda $$ the logarithmic natural eigenvalue and 
define $NNN_\lambda(x_i)$ as the logarithmic natural nearest neighborhood of $x_i$, denoted by $NNN(x_i)$, for $x_i\in X$.
\end{definition}

By Definition~\ref{LNNN},  the logarithmic natural eigenvalue $\lambda$ represents the number of iterations in the process of searching the logarithmic natural nearest neighborhood $NNN(X)$, which will be bounded by $\ln n+\ln\lambda$. The logarithmic natural nearest neighborhood $NNN(X)$ may contain an empty set.  
The robust (logarithmic) natural nearest neighborhood searching algorithm proposed in this paper, which is based on the NNN searching algorithm \cite{ZHFZ2014},  is shown in \myalgref{NNN-algorithm}.
\begin{algorithm}
	\caption{(Logarithmic) Natural Nearest Neighborhood Search Algorithm (NNN)}
	\label{NNN-algorithm}
	\begin{algorithmic}[1]
		\REQUIRE A set of points $X=\{x_1, x_2,...,x_n\} \subset \mathbb{R}^d$
		\ENSURE The (logarithmic) natural eigenvalue $\lambda=r$ and $NNN(x_i)=NNN_r(x_i)$ for every $x_i\in X$ 
		\STATE $\forall x_i \in X$, initially set $NN_0(x_i)=\emptyset$, $RNN_0(x_i)=\emptyset$, $NNN_0(x_i)=\emptyset$
		\STATE $r=1$, flag=0, $T=0$
		\WHILE{flag=0}
		\FOR{$\forall x_i \in X$}
		\STATE calculate $nn_r(x_i)=x_j$
		\STATE Reset $NN_r(x_i) =NN_{r-1}(x_i) \cup \{x_j\}$
		\STATE Reset $RNN_r(x_j) = RNN_{r-1}(x_j) \cup\{x_i\}$
		\STATE Reset $NNN_r(x_i) = NN_r(x_i) \cap RNN_r(x_i)$
		\ENDFOR
		\STATE calculate the set $NNN_{r}^0(X)$
		\IF{$NNN_{r}^0(X)$ remains unchanged}
		\STATE $T:=T+1$
		\ENDIF
		\IF{$T \geq \ln r+\ln n$ or $\emptyset \notin NNN_r(X)$}
		\STATE flag=1
		\ELSE
		\STATE $r:=r+1$
		\ENDIF
		\ENDWHILE
	\end{algorithmic}
\end{algorithm}

Remark: The NNN searching algorithm proposed by Zhu et al~\cite{ZHFZ2014} does not take outliers or noise into consideration. In contrast, by introducing $\ln n + \ln \lambda$ as the threshold, the robust natural nearest neighborhood searching algorithm we propose can recognize outliers and eliminate their effect.

\subsection{Local density estimation and selection of cluster centers} \label{selection-of-cluster-centers}
As shown in Section \ref{NNNsection}, NNN is a scale-free concept and does not require parameters in the selection of neighbors. The parameter $d_c$ in DPC can be avoided by changing the definition of the local density based on the NNN. For any data point $x_i$, the new local density is defined as
\begin{equation} \label{NNN-density}
	\rho_i = \sum_{x_j \in NNN(x_i)} \exp \left[-{\left( \frac{d_{ij}}{\sigma_i} \right) } ^2 \right],
\end{equation}
where $\sigma_i=\max\{d(x_i, x_j)\, |\, x_j\in NNN(x_i)\}$.
% is the maximum distance between $x_i$ and the points in $NNN(x_i)$. 
We define 
$$	\delta_i = \min \limits_{j : \rho_j > \rho_i} \left( d_{ij} \right)$$
and for data point $i$ which has the highest local density $\rho_i$, define
$$	\delta_i = \max \limits_j \left( d_{ij} \right), $$
which are the same as in DPC.

Let $\gamma_i=\rho_i\times \delta_i$ as defined in \myeqref{gamma}. We can choose the candidate centers based on a standard normal distribution with a one-sided $95\%$ confidence interval. The candidate centers are defined as
\begin{equation} \label{candidate-centers}
	CanCens = \{ x_i \in X\,  |\, \gamma_i > E(\gamma) + 1.65\var(\gamma) \},
\end{equation}
where $E(\gamma)$ is the expectation of $\gamma$, $\var(\gamma)$ is the variance of $\gamma$, 1.65 is the $95\%$ quantile of the standard normal distribution.

By the definition of CanCens, it does not contain the data points with low local density $\rho_i$ and small distance $\delta_i$, but it may have some points with large $\rho$ and small $\delta$ or small $\rho$ and large $\delta$ which is not suitable for being cluster centers. To eliminate those data points, we define
\begin{equation} \label{theta}
	\theta_i = \frac{\rho_i}{\delta_i}
\end{equation}
for every $x_i\in Cancens$. Note that the gap between $\rho_i$ and $\delta_i$ is large enough for those data points with large $\rho$ and small $\delta$ or small $\rho$ and large $\delta$. Thersfore, we can select the final centers from Cancens based on standard normal distribution of $\theta$ with a two-sided $95\%$ confidence interval as shown in \myeqref{final-centers}.
\begin{equation} \label{final-centers}
	\begin{split}
		Cens(X)=\{ x_i \in CanCens\, |\, \left|\theta_i-E(\theta)\right| <1.96{\var(\theta)}\},  
	\end{split}
\end{equation}
where 1.96 is the $97.5\%$ quantile of the standard normal distribution.

In the Probabilistic Propagation Clustering Algorithm provided in Section~\ref{PPsection}, one cluster may have more than one center point in $Cens(X)$,
% is usually larger than the accurate cluster number, 
so the allocation strategy of DPC cannot be used here. We need a new allocation strategy to cluster all the data points using the centers in $Cens(X)$.

\subsection{Probabilistic Propagation Clustering Algorithm} \label{PPsection}
The idea of the probabilistic propagation clustering algorithm is based on the spread of the epidemic in recent years. An infected man will infect the persons he contacts. Therefore, we first choose the data point with the largest local density $\rho$ in $Cens(X)$ as patient zero. In addition, he will infect the persons he contacts with, that is his natural nearest neighborhood ($NNN$). These neighbors will infect their natural nearest neighbors and the propagation will continue until there is no neighbor that can be infected. At this time, we recognize all the infected data points as a cluster. We call the processes of forming a cluster a round of propagations. Second, we select the data point that has the largest local density among those data points not infected in $Cens(X)$, and repeat the process above until all the data points in $Cens$ have been infected. 

In the real world, some people will not become infected even when they contact with infected persons because they have antibodies or other reasons. Therefore, in the algorithm, a data point $x\in X$ is infected with a certain probability  $p_x = C\cdot(p_{x}' + p_{x}'')$,  {where $p_x'$ (and $p_x''$) denotes the rank of $x$ in the current round of propagation. Formally, in the current round of propagation, if point $x$ (to be checked) and the infected points can be ordered $y_1,\ldots, y_{i-1}, x, y_{i+1},\ldots, y_k$ with $\rho(y_1)\le\ldots\le \rho(y_{i-1})\le\rho(x)\le\rho(y_{i+1})\le\ldots\le\rho(y_k)$, then we define 
$$p_x'=\frac{i}{k};$$
if point $x$ and the unchecked points in $\cup_{j=1, j\not=i}^{k} NNN(y_j)$ can be ordered $z_1,\ldots, z_{j-1}, x, z_{j+1},\ldots, z_\ell$ with $\rho(z_1)\le\ldots\le \rho(z_{j-1})\le\rho(x)\le\rho(z_{j+1})\le\ldots\le\rho(z_\ell)$, then we define 
$$p_x''=\frac{j}{\ell}.$$}

If point $x$ is not infected, we consider it having antibodies or dead and that it will no longer be infected. At the beginning of the propagation process, we increase the probability $p_x$ of being infected to ensure that the dissemination process will continue. At the end of the propagation process, we decrease the probability $p_x$ of being infected to guarantee that the outliers or noise will not be classified into any cluster.

\begin{algorithm}[ht] 
	\caption{Probabilistic propagation clustering algorithm  (PP)}
	\label{PPNNN-algorithm}
	\begin{algorithmic}[1]    % [1]表示所有代码前面加序号
		\REQUIRE A set of points $X \subset \mathbb{R}^d$, $\rho=\rho(X)$, $NNN(X)$
		\ENSURE Clustering results $CLU$
		\STATE Initially set  $Cens=Cens(X)$ 
		\STATE $n=1$
		\STATE $x=find(\max\rho(Cens)); Cens = Cens\setminus\{x\};$\\$CLU(n)=\emptyset,;CLU_{can}=NNN(x)$;
		\WHILE{$CLU_{can} \neq \emptyset$}
		\FOR{$\forall y \in CLU_{can}$}
		\STATE calculate the probability of being infected for y: $p_y$
		\IF {$y$ is infected (with probablity $p_y$)}
		\STATE Reset $CLU(n) = CLU(n) \cup \{y\}$	
		\STATE Reset $CLU_{can} = CLU_{can} \cup NNN(y)$
		\IF{$y \in Cens$}
		\STATE Reset $Cens = Cens\setminus\{y\}$
		\ENDIF
		\ELSE
		\STATE y does not belong to any cluster
		\ENDIF
		\STATE Reset $CLU_{can} = CLU_{can} \setminus \{y\}$
		\ENDFOR
		\ENDWHILE
		\IF{$Cens \neq \emptyset$}
		\STATE $n=n+1$
		\STATE goto Step 3
		\ENDIF
		\STATE Assign each remaining data point to the cluster that has the largest sum of local densities among its natural nearest neighborhood
	\end{algorithmic}
\end{algorithm}

When the propagation process is over, there will be some data points that are not infected and are not outliers. We assign each remaining data point to the cluster that has the largest sum of local densities among the natural nearest neighborhoods containing it.

By introducing the idea of probabilistic propagation, we can make the cluster algorithm closer to reality. Moreover, we can distinguish two clusters that are close to each other. The probabilistic propagation clustering algorithm is shown in \myalgref{PPNNN-algorithm}.
Denote $\rho(A)=\{\rho(x)\, |\, x\in A\}$.

\section{Experiment} \label{experiment-section}
{We call the algorithms~\ref{NNN-algorithm} and~\ref{PPNNN-algorithm} together the DPC-PPNNN algorithm.} To test the performance of the DPC-PPNNN algorithm, we use classical synthetic datasets and real-world datasets. Moreover, we take traditional DPC, K-means and DBSCAN as the control group, where the K-means and DBSCAN algorithms are implemented in the sklearn library of Python and the DPC algorithm, without the “Halo” part, is based on the source code provided by the author since our datasets do not contain noise. All the results shown are the optimal results after argument tuning.

The synthetic datasets and real-world datasets used in the experiments are presented in \mytabref{synthetic} and \mytabref{real-world}, respectively.
\begin{table*}[htbp]  % 人工数据集详细信息
	\centering
	\caption{Synthetic datasets.}
	\label{synthetic}
	\begin{tabular}{cccc}
		\toprule
		\textbf{Dataset} & \textbf{No. of records} & \textbf{No. of attributes} & \textbf{No. of clusters} \\ \midrule
		2d-4c-no9 & 876 & 2 & 4 \\  
		3-spiral & 312 & 2 & 3 \\ 
		Aggregation & 788 & 2 & 7 \\ 
		Cassini & 1000 & 2 & 3 \\
		Complex9 & 3031 & 2 & 9 \\   
		Compound & 399 & 2 & 6 \\ 
		Dartboard1 & 1000 & 2 & 4 \\ 
		Jain & 373 & 2 & 2 \\ 
		R15 & 600 & 2 & 15 \\ 
		Shapes & 1000 & 2 & 4 \\
		\bottomrule
	\end{tabular}
\end{table*}
\begin{table*}[htbp]   % 真实数据集详细信息
	\centering
	\caption{Real-world datasets.}
	\label{real-world}
	\begin{tabular}{cccc}
		\toprule
		\textbf{Dataset} & \textbf{No. of records} & \textbf{No. of attributes} & \textbf{No. of clusters} \\ \midrule
		Ecoli & 336 & 7 & 8 \\ 
		Glass & 214 & 9 & 7 \\
		Heart-statlog & 270 & 13 & 2 \\
		Iono & 351 & 34 & 2 \\
		Iris & 150 & 4 & 3 \\
		Thy & 215 & 5 & 3 \\
		Wdbc & 569 & 30 & 2 \\
		Wine & 178 & 13 & 3 \\
		%        Olivetti faces & 400 & $92 \times 112$ & 40 \\
		\bottomrule
	\end{tabular}
\end{table*}
\begin{table*}[htbp]     % 人工数据集的聚类结果表
	\centering
	\caption{Performances of different clustering algorithms on different synthetic datasets.}
	\label{synthetic-results}
	\begin{tabular}{llllllll}
		\toprule
		\textbf{Algorithm} & \textbf{ARI} & \textbf{AMI} & \textbf{FMI} &  & \textbf{ARI} & \textbf{AMI} & \textbf{FMI} \\ \midrule
		& \multicolumn{2}{l}{2d-4c-no9} &  &  & \multicolumn{2}{l}{Compound} & \\
		DPC-PPNNN(ours) & \textbf{0.9931} & \textbf{0.9898} & \textbf{0.9951} &  & \textbf{0.9375} & \textbf{0.9345} & \textbf{0.9543} \\ 
		DPC\cite{RL2014} & 0.3336 & 0.5261 & 0.6108 &  & 0.5099 & 0.7542 & 0.6235 \\
		K-means\cite{MJ1967} & 0.8952 & 0.9007 & 0.9241 &  & 0.5364 & 0.7128 & 0.6410 \\
		DBSCAN\cite{EKHSX1996} & 0.8412 & 0.8810 & 0.8940 &  & 0.8774 & 0.8673 & 0.9103 \\
		& \multicolumn{2}{l}{3-spiral} &  &  & \multicolumn{2}{l}{Dartboard1} & \\
		DPC-PPNNN(ours) & \textbf{1.0000} & \textbf{1.0000} & \textbf{1.0000} &  & \textbf{1.0000} & \textbf{1.0000} & \textbf{1.0000} \\ 
		DPC\cite{RL2014} & 0.9521 & 0.9406 & 0.9680 &  & 0.0006 & 0.0130 & 0.3213 \\
		K-means\cite{MJ1967} & -0.0060 & -0.0055 & 0.3274 &  & -0.0030 & -0.0033 & 0.2470 \\
		DBSCAN\cite{EKHSX1996} & 0.9953 & 0.9918 & 0.9969 &  & \textbf{1.0000} & \textbf{1.0000} & \textbf{1.0000} \\
		& \multicolumn{2}{l}{Aggregation} &  &  & \multicolumn{2}{l}{Jain} & \\
		DPC-PPNNN(ours) & 0.9927 & 0.9881 & 0.9943 &  & \textbf{1.0000} & \textbf{1.0000} & \textbf{1.0000} \\ 
		DPC\cite{RL2014} & \textbf{0.9978} & \textbf{0.9956} & \textbf{0.9983} &  & 0.6438 & 0.5951 & 0.8502 \\
		K-means\cite{MJ1967} & 0.7624 & 0.8776 & 0.8159 &  & 0.3241 & 0.3677 & 0.7005 \\
		DBSCAN\cite{EKHSX1996} & 0.8719 & 0.9223 & 0.9051 &  & 0.9350 & 0.8476 & 0.9742 \\
		& \multicolumn{2}{l}{Cassini} &  &  & \multicolumn{2}{l}{R15} & \\
		DPC-PPNNN(ours) & \textbf{1.0000} & \textbf{1.0000} & \textbf{1.0000} &  & \textbf{0.9928} & \textbf{0.9938} & \textbf{0.9932} \\ 
		DPC\cite{RL2014} & 0.4886 & 0.5879 & 0.6736 &  & 0.9857 & 0.9885 & 0.9866 \\
		K-means\cite{MJ1967} & 0.5308 & 0.5400 & 0.7012 &  & \textbf{0.9928} & \textbf{0.9938} & \textbf{0.9932} \\
		DBSCAN\cite{EKHSX1996} & \textbf{1.0000} & \textbf{1.0000} & \textbf{1.0000} &  & 0.8307 & 0.8905 & 0.8416 \\
		& \multicolumn{2}{l}{Complex9} &  &  & \multicolumn{2}{l}{Shapes} & \\
		DPC-PPNNN(ours) & 0.9375 & 0.9345 & 0.9543 &  & \textbf{1.0000} & \textbf{1.0000} & \textbf{1.0000} \\ 
		DPC\cite{RL2014} & 0.3929 & 0.6666 & 0.4940 &  & 0.7439 & 0.8410 & 0.8145 \\
		K-means\cite{MJ1967} & 0.3461 & 0.6161 & 0.4517 &  & \textbf{1.0000} & \textbf{1.0000} & \textbf{1.0000} \\
		DBSCAN\cite{EKHSX1996} & \textbf{0.9998} & \textbf{0.9994} & \textbf{0.9998} &  & \textbf{1.0000} & \textbf{1.0000} & \textbf{1.0000} \\
		\bottomrule
	\end{tabular}
\end{table*}
\begin{figure*}[htbp] % 插入图片三个反例应用新算法的正确结果
	\centering
	\subfigure[counter example 1]{
		\begin{minipage}[t]{0.3\linewidth}
			\centering
			\includegraphics[width=1.5in]{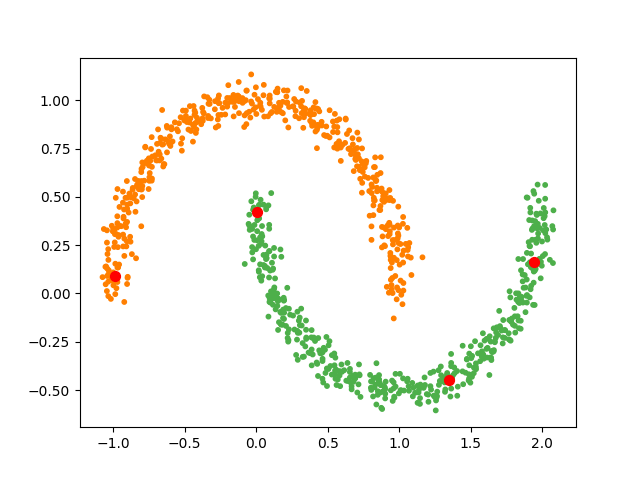}
	\end{minipage}}
	\subfigure[counter example 2]{
		\begin{minipage}[t]{0.3\linewidth}
			\centering
			\includegraphics[width=1.5in]{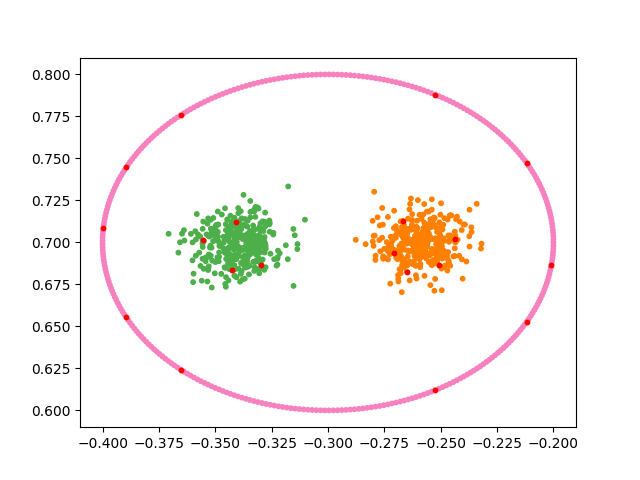}
	\end{minipage}}
	\subfigure[counter example 3]{
		\begin{minipage}[t]{0.3\linewidth}
			\centering
			\includegraphics[width=1.5in]{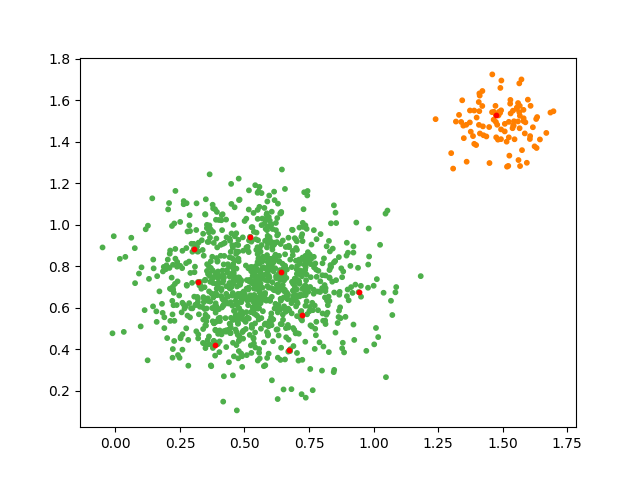}
	\end{minipage}}
	\caption{The re-clustering results of the three counter examples in Section \ref{analysis}}
	\label{3-right-samples}
\end{figure*}
\begin{figure*}[htbp] % 插入2d-4c-no9的四个结果
	\centering
	\subfigure[DPC-PPNNN on 2d-4c-no9]{
		\begin{minipage}[l]{0.2\linewidth}
			\centering
			\includegraphics[width=1.5in]{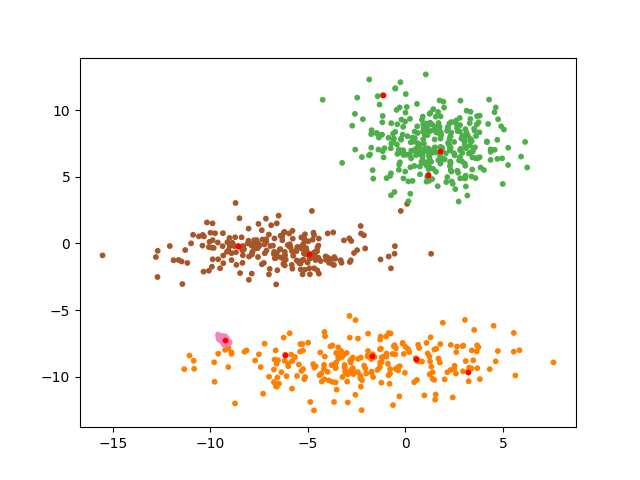}
	\end{minipage}}
	\subfigure[DPC on 2d-4c-no9]{
		\begin{minipage}[l]{0.2\linewidth}
			\centering
			\includegraphics[width=1.5in]{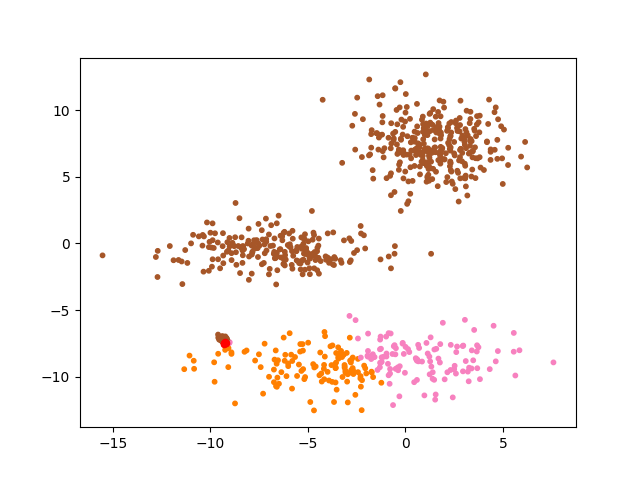}
	\end{minipage}}
	\subfigure[K-means on 2d-4c-no9]{
		\begin{minipage}[l]{0.2\linewidth}
			\centering
			\includegraphics[width=1.5in]{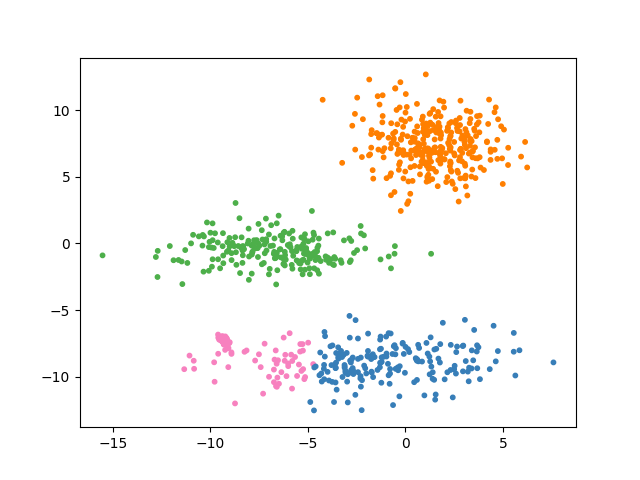}
	\end{minipage}}
	\subfigure[DBSCAN on 2d-4c-no9]{
		\begin{minipage}[l]{0.2\linewidth}
			\centering
			\includegraphics[width=1.5in]{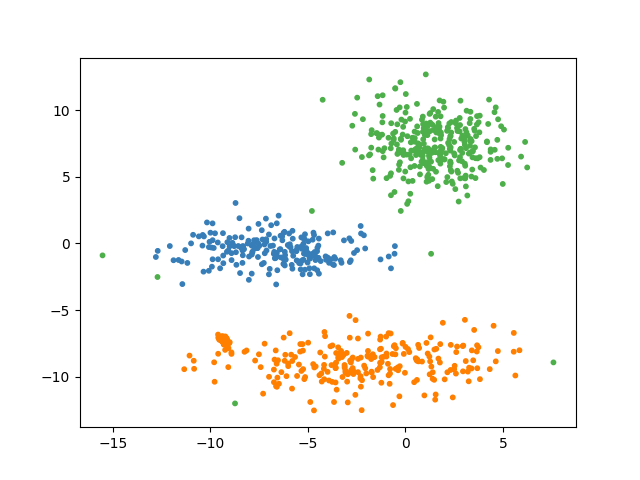}
	\end{minipage}}
	\caption{The clustering results on 2d-4c-no9 by 4 algorithms.}
	\label{2d-4c-no9}
\end{figure*}
\begin{figure*}[htbp] % 插入3-spiral的四个结果
	\centering
	\subfigure[DPC-PPNNN on 3-spiral]{
		\begin{minipage}[t]{0.2\linewidth}
			\centering
			\includegraphics[width=1.5in]{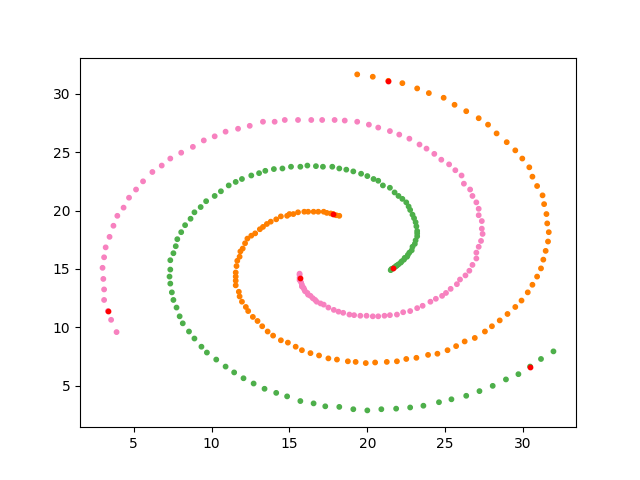}
	\end{minipage}}
	\subfigure[DPC on 3-spiral]{
		\begin{minipage}[t]{0.2\linewidth}
			\centering
			\includegraphics[width=1.5in]{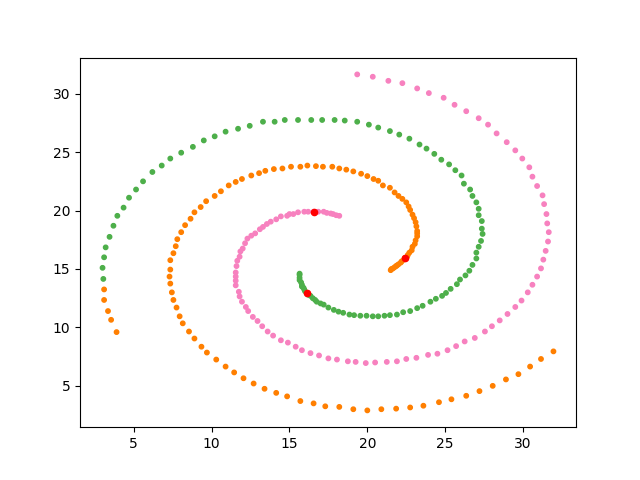}
	\end{minipage}}
	\subfigure[K-means on 3-spiral]{
		\begin{minipage}[t]{0.2\linewidth}
			\centering
			\includegraphics[width=1.5in]{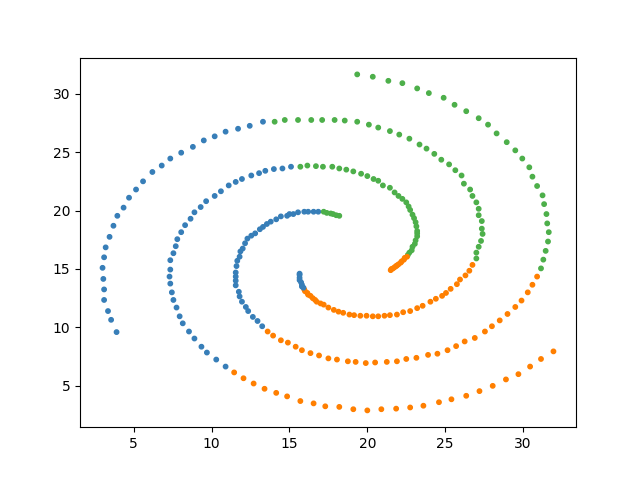}
	\end{minipage}}
	\subfigure[DBSCAN on 3-spiral]{
		\begin{minipage}[t]{0.2\linewidth}
			\centering
			\includegraphics[width=1.5in]{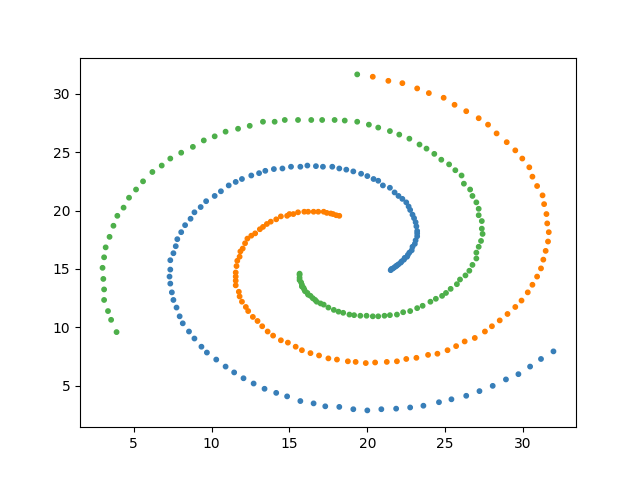}
	\end{minipage}}
	\caption{The clustering results on 3-spiral by 4 algorithms.}
	\label{3-spiral}
\end{figure*}
\begin{figure*}[htbp] % 插入aggregation的四个结果
	\centering
	\subfigure[DPC-PPNNN on Aggregation]{
		\begin{minipage}[t]{0.2\linewidth}
			\centering
			\includegraphics[width=1.5in]{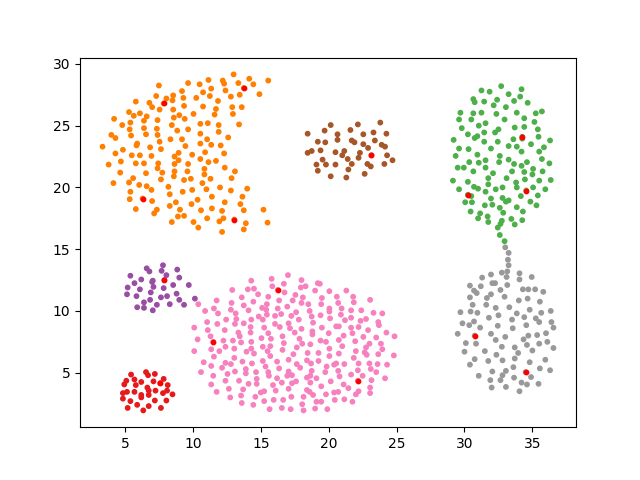}
	\end{minipage}}
	\subfigure[DPC on Aggregation]{
		\begin{minipage}[t]{0.2\linewidth}
			\centering
			\includegraphics[width=1.5in]{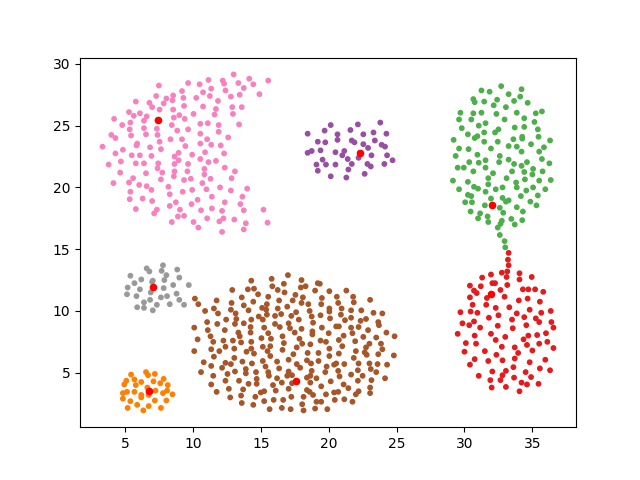}
	\end{minipage}}
	\subfigure[K-means on Aggregation]{
		\begin{minipage}[t]{0.2\linewidth}
			\centering
			\includegraphics[width=1.5in]{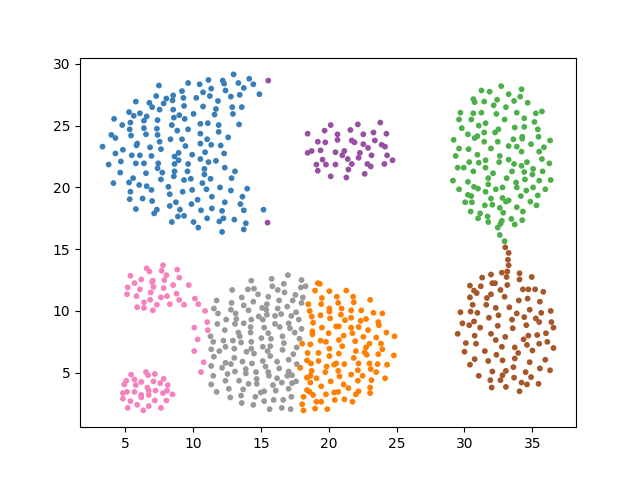}
	\end{minipage}}
	\subfigure[DBSCAN on Aggregation]{
		\begin{minipage}[t]{0.2\linewidth}
			\centering
			\includegraphics[width=1.5in]{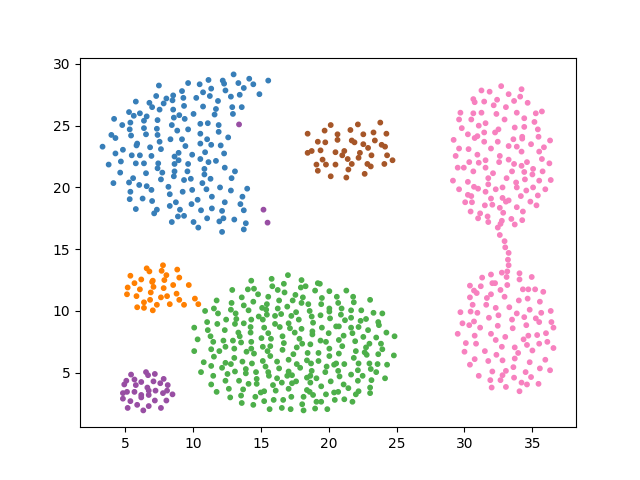}
	\end{minipage}}
	\caption{The clustering results on Aggregation by 4 algorithms.}
	\label{Aggregation}
\end{figure*}
\begin{figure*}[htbp] % 插入cassini的四个结果
	\centering
	\subfigure[DPC-PPNNN on Cassini]{
		\begin{minipage}[t]{0.2\linewidth}
			\centering
			\includegraphics[width=1.5in]{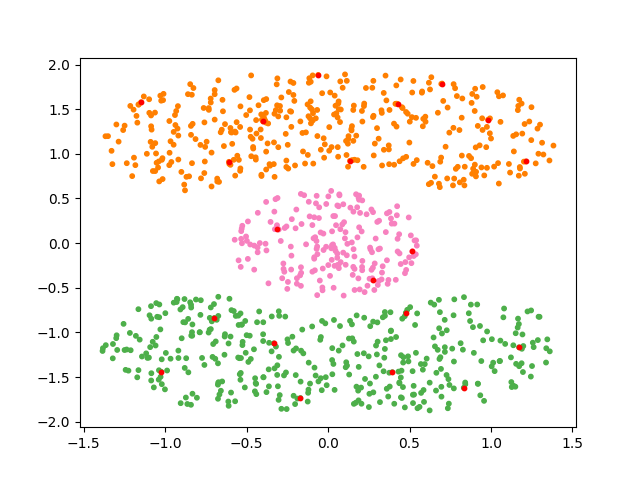}
	\end{minipage}}
	\subfigure[DPC on Cassini]{
		\begin{minipage}[t]{0.2\linewidth}
			\centering
			\includegraphics[width=1.5in]{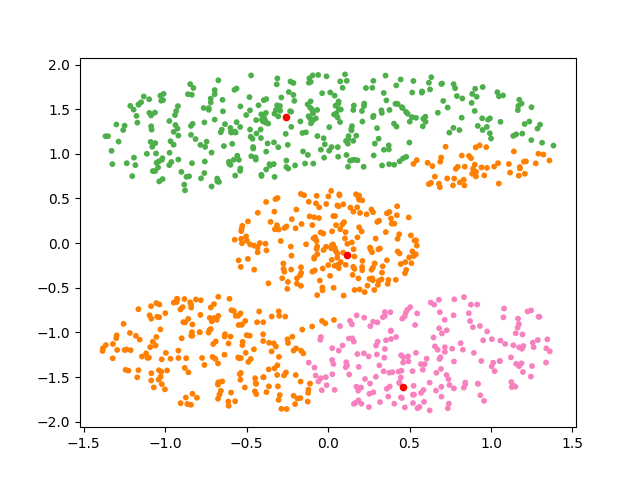}
	\end{minipage}}
	\subfigure[K-means on Cassini]{
		\begin{minipage}[t]{0.2\linewidth}
			\centering
			\includegraphics[width=1.5in]{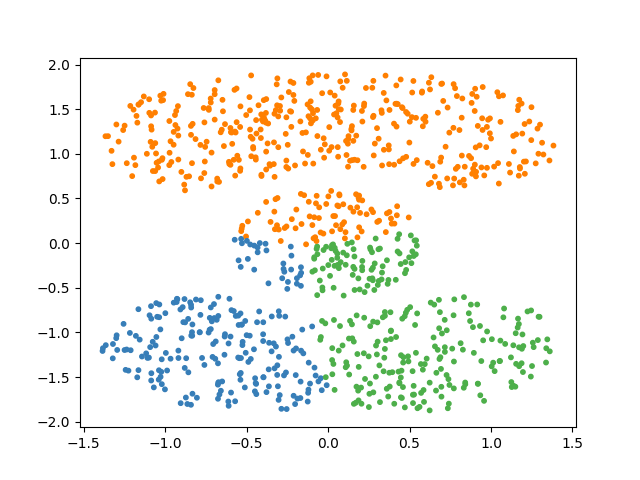}
	\end{minipage}}
	\subfigure[DBSCAN on Cassini]{
		\begin{minipage}[t]{0.2\linewidth}
			\centering
			\includegraphics[width=1.5in]{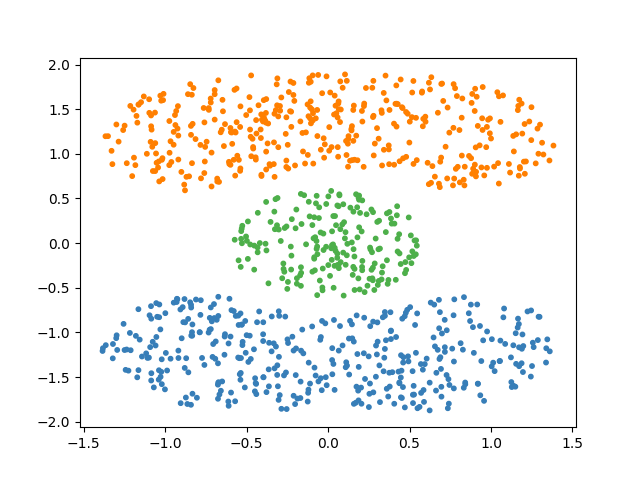}
	\end{minipage}}
	\caption{The clustering results on Cassini by 4 algorithms.}
	\label{Cassini}
\end{figure*}
\begin{figure*}[htbp] % 插入complex9的四个结果
	\centering
	\subfigure[DPC-PPNNN on Complex9]{
		\begin{minipage}[t]{0.2\linewidth}
			\centering
			\includegraphics[width=1.5in]{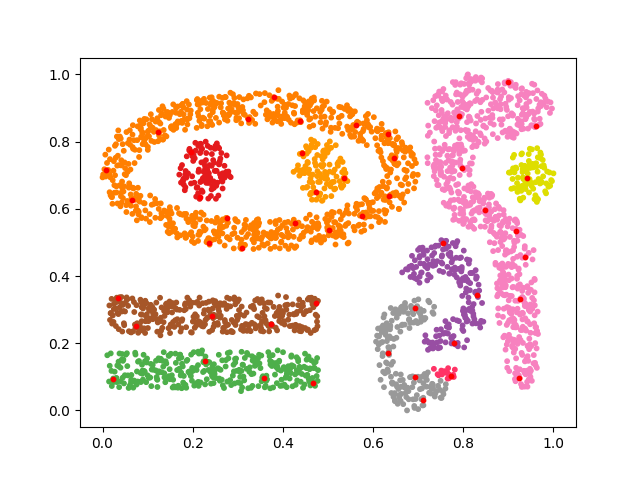}
	\end{minipage}}
	\subfigure[DPC on Complex9]{
		\begin{minipage}[t]{0.2\linewidth}
			\centering
			\includegraphics[width=1.5in]{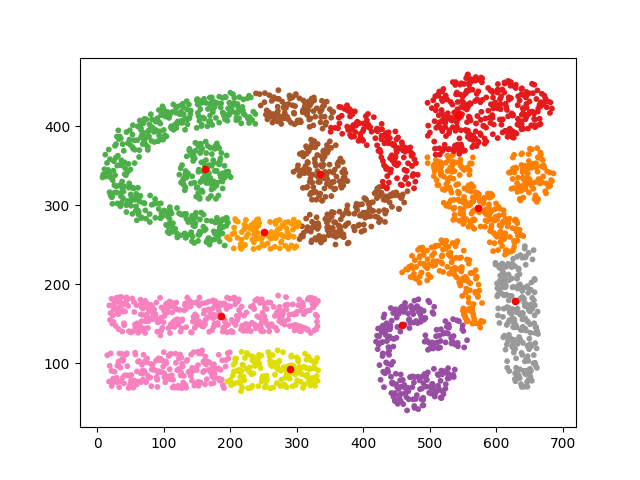}
	\end{minipage}}
	\subfigure[K-means on Complex9]{
		\begin{minipage}[t]{0.2\linewidth}
			\centering
			\includegraphics[width=1.5in]{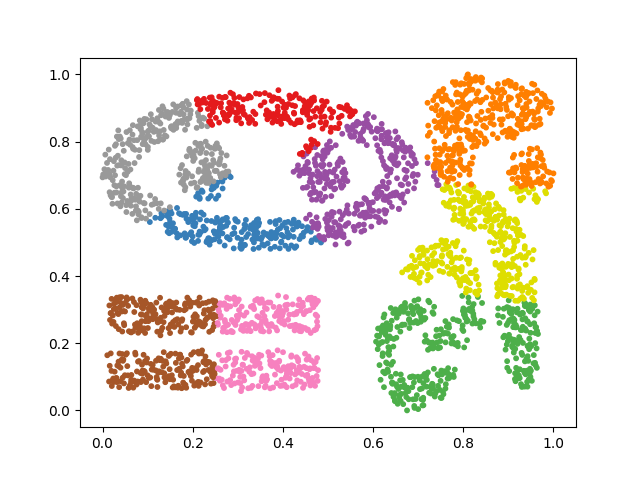}
	\end{minipage}}
	\subfigure[DBSCAN on Complex9]{
		\begin{minipage}[t]{0.2\linewidth}
			\centering
			\includegraphics[width=1.5in]{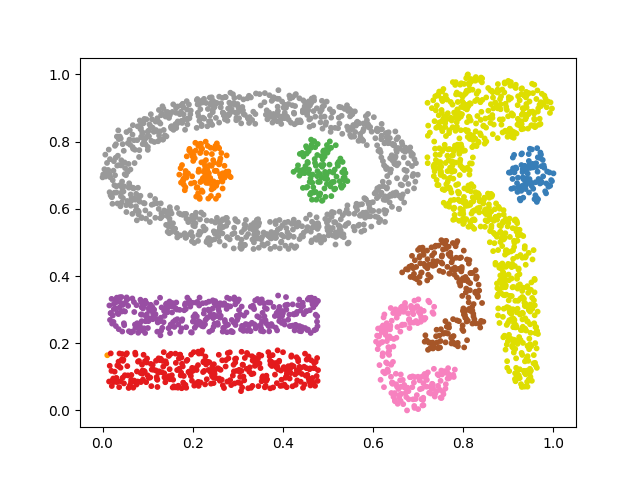}
	\end{minipage}}
	\caption{The clustering results on Complex9 by 4 algorithms.}
	\label{Complex9}
\end{figure*}
\begin{figure*}[htbp] % 插入compound的四个结果
	\centering
	\subfigure[DPC-PPNNN on Compound]{
		\begin{minipage}[t]{0.2\linewidth}
			\centering
			\includegraphics[width=1.5in]{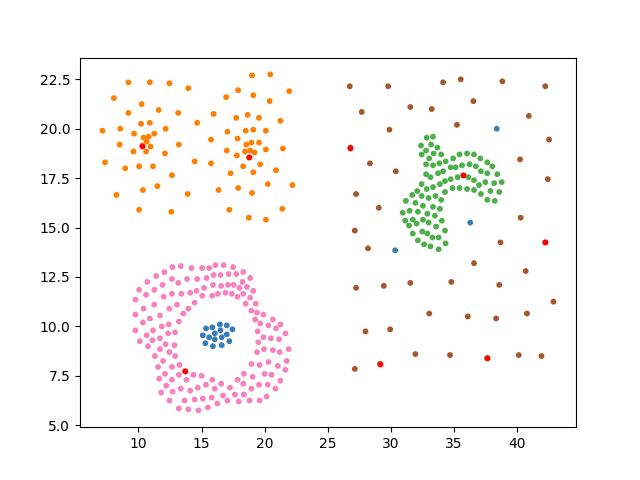}
	\end{minipage}}
	\subfigure[DPC on Compound]{
		\begin{minipage}[t]{0.2\linewidth}
			\centering
			\includegraphics[width=1.5in]{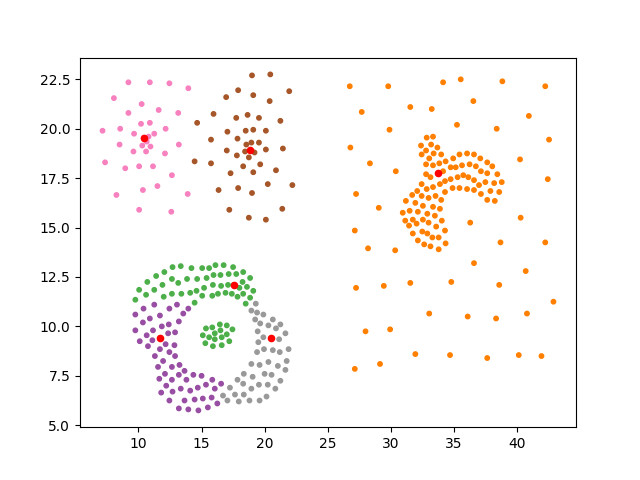}
	\end{minipage}}
	\subfigure[K-means on Compound]{
		\begin{minipage}[t]{0.2\linewidth}
			\centering
			\includegraphics[width=1.5in]{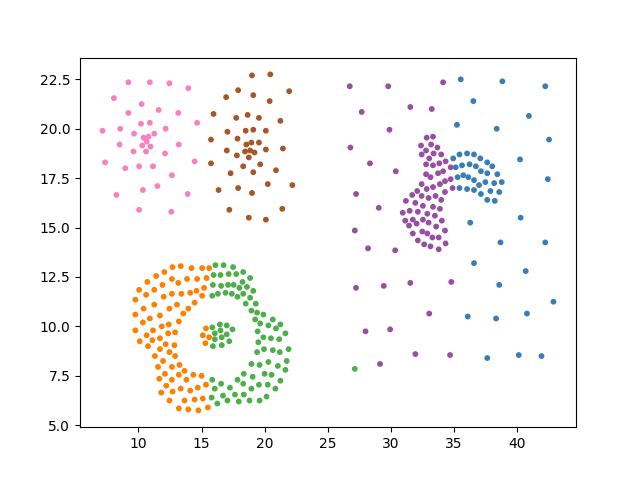}
	\end{minipage}}
	\subfigure[DBSCAN on Compound]{
		\begin{minipage}[t]{0.2\linewidth}
			\centering
			\includegraphics[width=1.5in]{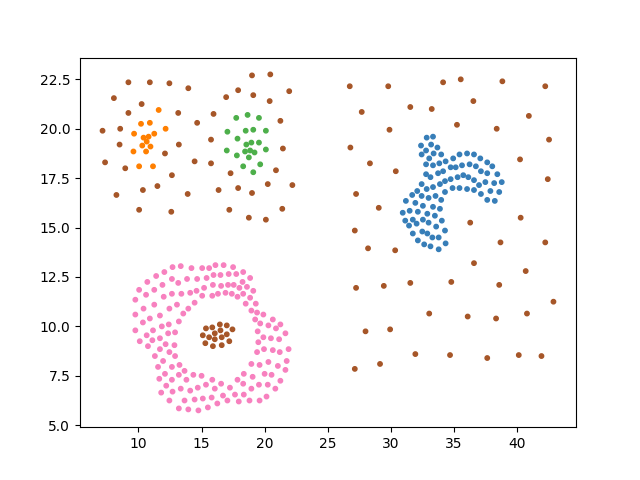}
	\end{minipage}}
	\caption{The clustering results on Compound by 4 algorithms.}
	\label{Compound}
\end{figure*}
\begin{figure*}[htbp] % 插入dartboard1的四个结果
	\centering
	\subfigure[DPC-PPNNN on Dartboard1]{
		\begin{minipage}[t]{0.2\linewidth}
			\centering
			\includegraphics[width=1.5in]{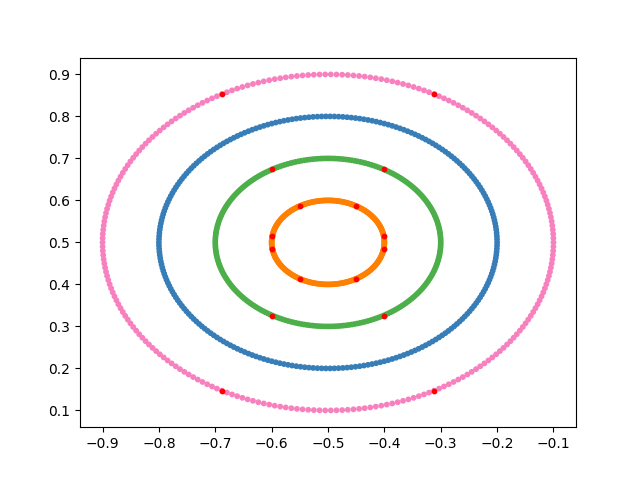}
	\end{minipage}}
	\subfigure[DPC on Dartboard1]{
		\begin{minipage}[t]{0.2\linewidth}
			\centering
			\includegraphics[width=1.5in]{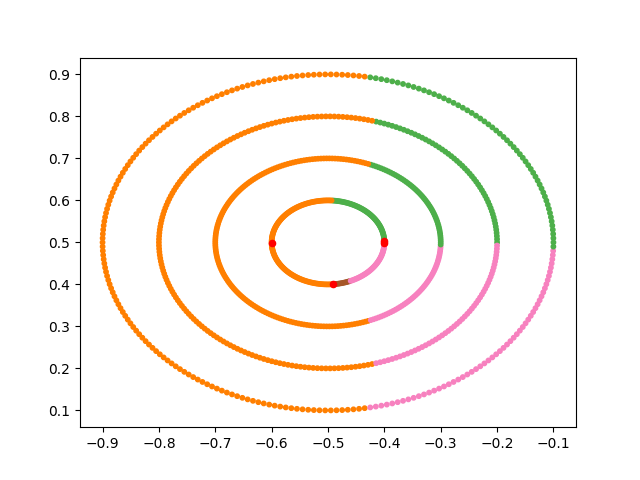}
	\end{minipage}}
	\subfigure[K-means on Dartboard1]{
		\begin{minipage}[t]{0.2\linewidth}
			\centering
			\includegraphics[width=1.5in]{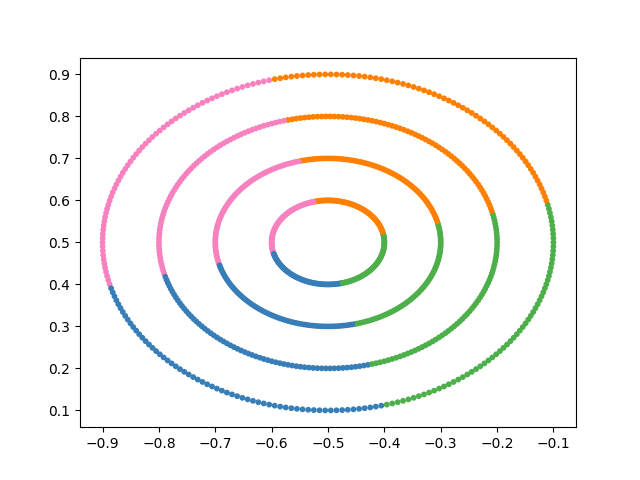}
	\end{minipage}}
	\subfigure[DBSCAN on Dartboard1]{
		\begin{minipage}[t]{0.2\linewidth}
			\centering
			\includegraphics[width=1.5in]{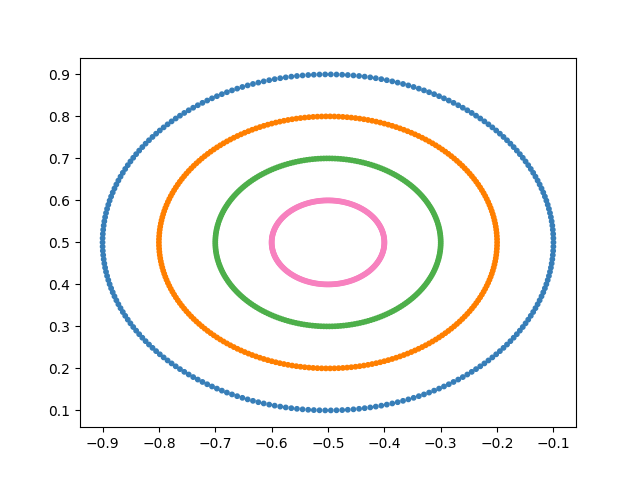}
	\end{minipage}}
	\caption{The clustering results on Dartboard1 by 4 algorithms.}
	\label{Dartboard1}
\end{figure*}
\begin{figure*}[htbp] % 插入jain的四个结果
	\centering
	\subfigure[DPC-PPNNN on Jain]{
		\begin{minipage}[t]{0.2\linewidth}
			\centering
			\includegraphics[width=1.5in]{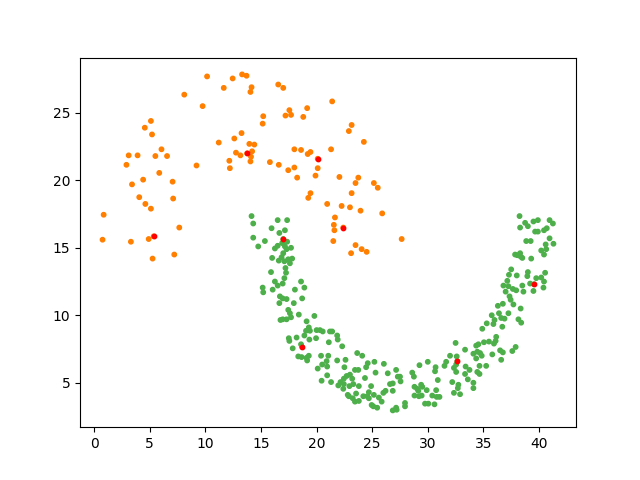}
	\end{minipage}}
	\subfigure[DPC on Jain]{
		\begin{minipage}[t]{0.2\linewidth}
			\centering
			\includegraphics[width=1.5in]{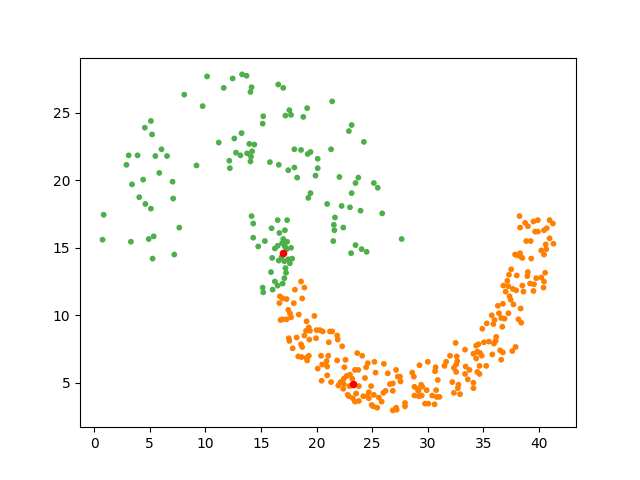}
	\end{minipage}}
	\subfigure[K-means on Jain]{
		\begin{minipage}[t]{0.2\linewidth}
			\centering
			\includegraphics[width=1.5in]{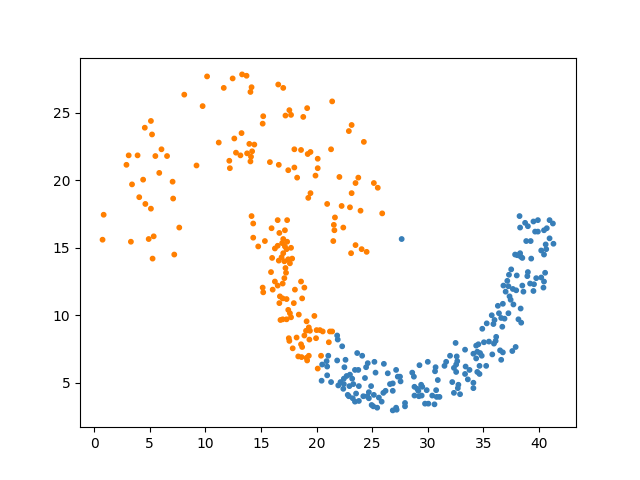}
	\end{minipage}}
	\subfigure[DBSCAN on Jain]{
		\begin{minipage}[t]{0.2\linewidth}
			\centering
			\includegraphics[width=1.5in]{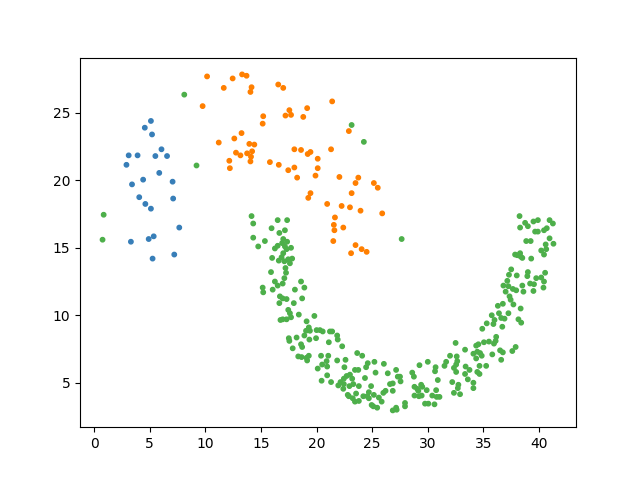}
	\end{minipage}}
	\caption{The clustering results on Jain by 4 algorithms.}
	\label{Jain}
\end{figure*}
\begin{figure*}[htbp] % 插入R15的四个结果
	\centering
	\subfigure[DPC-PPNNN on R15]{
		\begin{minipage}[t]{0.2\linewidth}
			\centering
			\includegraphics[width=1.5in]{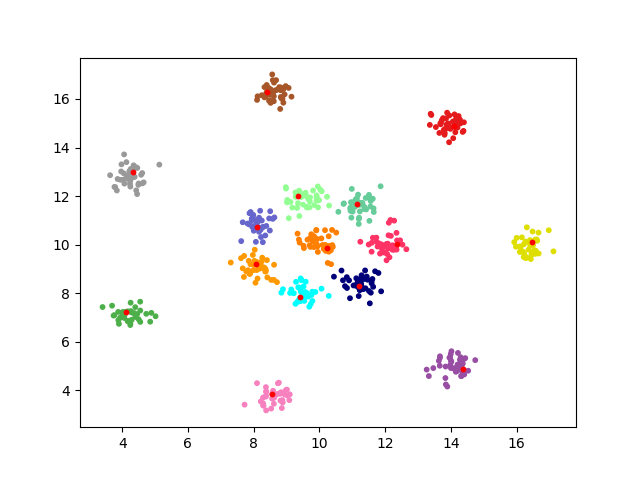}
	\end{minipage}}
	\subfigure[DPC on R15]{
		\begin{minipage}[t]{0.2\linewidth}
			\centering
			\includegraphics[width=1.5in]{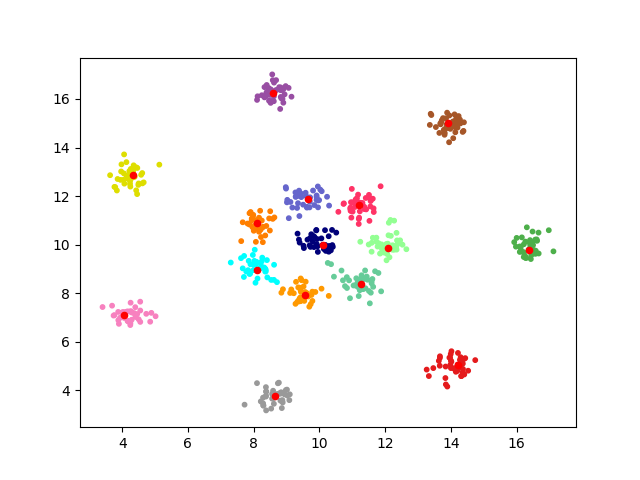}
	\end{minipage}}
	\subfigure[K-means on R15]{
		\begin{minipage}[t]{0.2\linewidth}
			\centering
			\includegraphics[width=1.5in]{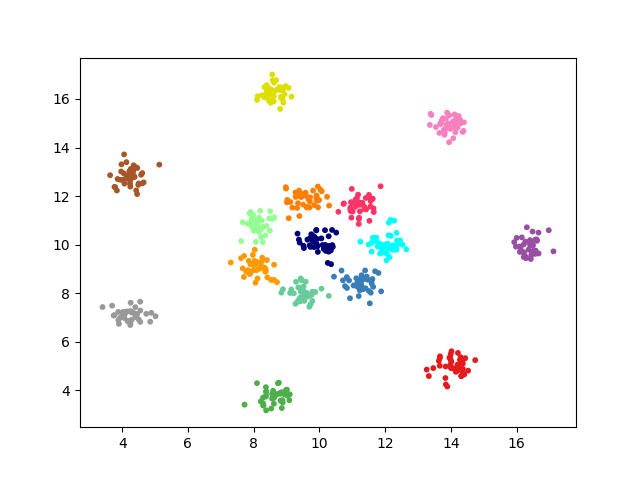}
	\end{minipage}}
	\subfigure[DBSCAN on R15]{
		\begin{minipage}[t]{0.2\linewidth}
			\centering
			\includegraphics[width=1.5in]{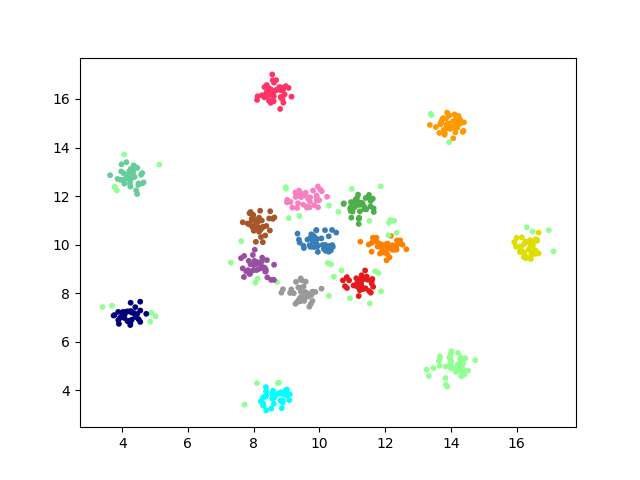}
	\end{minipage}}
	\caption{The clustering results on R15 by 4 algorithms.}
	\label{R15}
\end{figure*}
\begin{figure*}[htbp] % 插入shapes的四个结果
	\centering
	\subfigure[DPC-PPNNN on Shapes]{
		\begin{minipage}[t]{0.2\linewidth}
			\centering
			\includegraphics[width=1.5in]{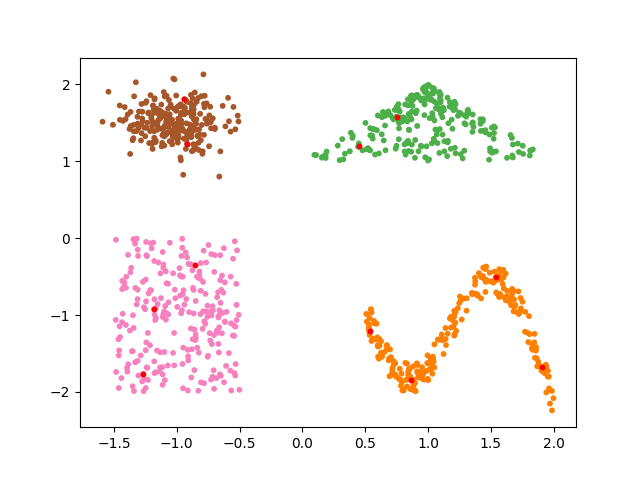}
	\end{minipage}}
	\subfigure[DPC on Shapes]{
		\begin{minipage}[t]{0.2\linewidth}
			\centering
			\includegraphics[width=1.5in]{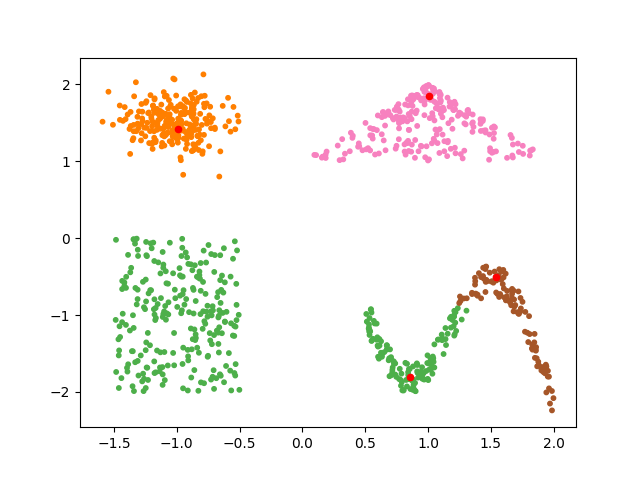}
	\end{minipage}}
	\subfigure[K-means on Shapes]{
		\begin{minipage}[t]{0.2\linewidth}
			\centering
			\includegraphics[width=1.5in]{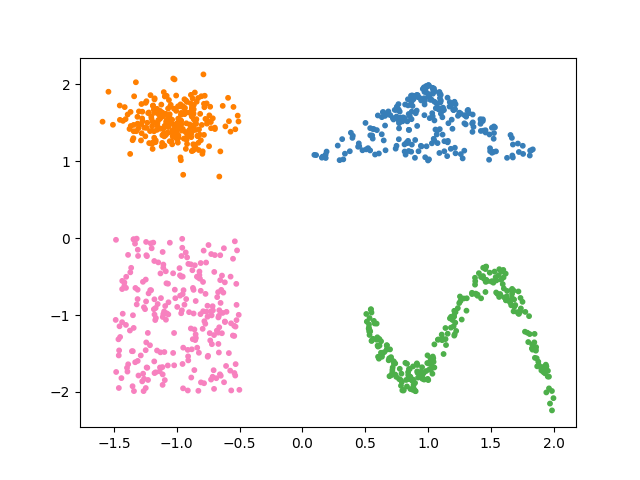}
	\end{minipage}}
	\subfigure[DBSCAN on Shapes]{
		\begin{minipage}[t]{0.2\linewidth}
			\centering
			\includegraphics[width=1.5in]{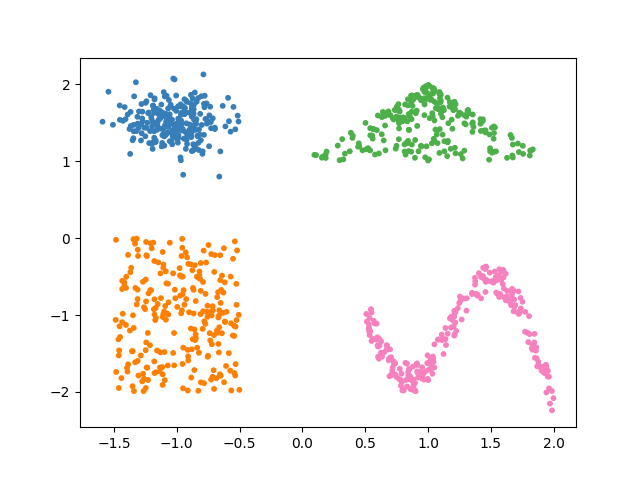}
	\end{minipage}}
	\caption{The clustering results on Shapes by 4 algorithms.}
	\label{Shapes}
\end{figure*}
\begin{table*}[htbp]   % 真实数据集的聚类结果
	\centering
	\caption{Performances of different clustering algorithms on different real-world datasets.}
	\label{real-world-results}
	\begin{tabular}{llllllll}
		\toprule
		\textbf{Algorithm} & \textbf{ARI} & \textbf{AMI} & \textbf{FMI} &  & \textbf{ARI} & \textbf{AMI} & \textbf{FMI} \\ \midrule
		& \multicolumn{2}{l}{Ecoli} &  &  & \multicolumn{2}{l}{Iris} & \\
		DPC-PPNNN(ours) & \textbf{0.5761} & 0.5614 & \textbf{0.7160} &  & \textbf{0.8680} & \textbf{0.8446} & \textbf{0.9114} \\
		DPC\cite{RL2014} & 0.2660 & 0.3179 & 0.4361 &  & 0.5414 & 0.6780 & 0.7539 \\
		K-means\cite{MJ1967} & 0.4115 & \textbf{0.5921} & 0.5482 &  & 0.7163 & 0.7387 & 0.8112 \\
		DBSCAN\cite{EKHSX1996} & 0.1004 & 0.1275 & 0.5282 &  & 0.5681 & 0.7316 & 0.7715 \\
		& \multicolumn{2}{l}{Glass} &  &  & \multicolumn{2}{l}{Thy} & \\
		DPC-PPNNN(ours) & \textbf{0.2595} & 0.3450 & \textbf{0.5534} &  & \textbf{0.7522} & \textbf{0.6108} & \textbf{0.8827} \\
		DPC\cite{RL2014} & 0.0651 & 0.0984 & 0.3516 &  & 0.2105 & 0.2301 & 0.7530 \\
		K-means\cite{MJ1967} & 0.1662 & 0.2942 & 0.3945 &  & 0.6283 & 0.5909 & 0.8546 \\
		DBSCAN\cite{EKHSX1996} & 0.2382 & \textbf{0.3577} & 0.5500 &  & 0.6715 & 0.4911 & 0.8606 \\
		& \multicolumn{2}{l}{Heart-statlog} &  &  & \multicolumn{2}{l}{Wdbc} & \\
		DPC-PPNNN(ours) & 0.2380 & 0.2186 & 0.5432 &  & 0.7193 & 0.6054 & 0.8671 \\
		DPC\cite{RL2014} & 0.0003 & 0.0002 & 0.7050 &  & 0.4705 & 0.4614 & 0.7860 \\
		K-means\cite{MJ1967} & \textbf{0.3488} & \textbf{0.2692} & \textbf{0.6755} &  & \textbf{0.7302} & \textbf{0.6226} & \textbf{0.8770} \\
		DBSCAN\cite{EKHSX1996} & 0.0861 & 0.1242 & 0.3417 &  & 0.2201 & 0.2618 & 0.6091 \\
		& \multicolumn{2}{l}{Iono} &  &  & \multicolumn{2}{l}{Wine} & \\
		DPC-PPNNN(ours) & 0.5287 & 0.4114 & 0.7935 &  & 0.7421 & 0.7268 & 0.8277 \\
		DPC\cite{RL2014} & 0.0019 & 0.0013 & 0.7302 &  & 0.6990 & 0.7233 & 0.8006 \\
		K-means\cite{MJ1967} & 0.1776 & 0.1330 & 0.6053 &  & \textbf{0.8537} & \textbf{0.8400} & \textbf{0.9026} \\
		DBSCAN\cite{EKHSX1996} & \textbf{0.7214} & \textbf{0.6343} & \textbf{0.8779} &  & 0.4229 & 0.5209 & 0.6482 \\
		\bottomrule
	\end{tabular}
\end{table*}

\subsection{Preliminaries}
To evaluate the performance of the clustering algorithm, three evaluation indices are introduced, namely, Adjusted Rand Index (ARI), Adjusted Mutual Information (AMI) and Fowlkes–Mallows index (FMI). The upper bound of the three indicators is 1, in other words, larger values of the indicators indicate better clustering results.

Before the tests, all of the real-world datasets should be adjusted by min–max normalization to eliminate the differences in the ranges of different dimensions, as shown in \myeqref{min-max}.
\begin{equation} \label{min-max}
	x^{'}_{ij} = \frac{x_{ij}-min(x_j)}{max(x_j)-min(x_j)}
\end{equation}
where 
%$x{'}_{ij}$ is the re-scaled data in the $i$-th position of $x_j$, 
$x_{ij}$ is the original data in the $i$-th position of $x_j$.
%, and $x_j$ is the original data in the entire $j$-th column.

To more objectively reflect the actual results of various algorithms, we perform argument tuning on each of the algorithms, thereby ensuring that their best performances are compared. For K-means, we simply give the correct number of clusters. For DBSCAN, we use a grid search to find the best parameter configuration. For traditional DPC, we choose the optimal $d_c$ from $1\%$ to $2\%$ and apply the Gaussian kernel in the density estimation process.

\subsection{Synthetic datasets}
In this part, we select a number of synthetic datasets that are widely used to test the performance of clustering algorithms. These datasets are different in terms of the distribution and numbers of points and clusters. They can simulate different situations to compare the performance of various clustering algorithms in different scenarios.

\mytabref{synthetic-results} shows the clustering results in terms of the ARI, AMI and FMI scores on all synthetic datasets listed in \mytabref{synthetic}. The evaluation criterion with the highest score is marked in bold, and we can see that in most cases, the DPC-PPNNN obtains  the highest score, except for the Complex9 (where the score of DPC-PPNNN is a bit less than the DBSCAN).

In \myfigref{3-right-samples}, we recluster the three counter examples in Section \ref{analysis} by DPC-PPNNN. Clearly, the results are much better than the original DPC which implies that our algorithm can overcome the deficiencies of DPC.

Next, we will show some typical clustering results on the synthetic datasets by DPC-PPNNN and the algorithms of the control group. The points with different colors in the figures are assigned to different clusters. The cluster centers obtained from DPC and DPC-PPNNN are marked in red.

In \myfigref{2d-4c-no9}, we can see that the 2d-4c-no9 dataset has four clusters, of which one cluster has a very high density. DPC-PPNNN succeeds in detecting all of them, while {the other algorithms of the control group fail to do so.

\myfigref{3-spiral} shows the results of each algorithm on the 3-spiral dataset. DPC-PPNNN can identify the four clusters correctly. Furthermore, DPC and DBSCAN detect the four clusters mostly correctly with some wrong data points at the tail, perhaps because the arguments are not set correctly, while KNN cannot recognize them at all.

The clustering results of the DPC-PPNNN and the other three algorithms of the control group on the Aggregation dataset are shown in \myfigref{Aggregation}. The dataset can be detected correctly by DPC-PPNNN and DPC. K-means fails to do so because the dataset is somewhat complex for K-means. DBSCAN cannot distinguish the two clusters on the right.

From the clustering results in \myfigref{Cassini}, we can see that DPC-PPNNN and DBSCAN can detect the clusters in the Cassini dataset. Although DPC can find the correct cluster centers, it fails to allocate the other remaining data points correctly. The three clusters are not uniform in shape which leads to the wrong cluster results by K-means.

The \myfigref{Complex9} shows the results of the four algorithms on the Complex9 dataset. DBSCAN can recognize all the clusters successfully. While for DPC-PPNNN, there are some misclassified points at the tail of a cluster because these points are not connected tightly enough. DPC finds the correct cluster centers but it fails to allocate the other remaining data points correctly. K-means has poor performance.

\myfigref{Compound} displays the results on the Compound dataset. DPC-PPNNN can detect most of the clusters correctly and can also recognize noise. DBSCAN is good at recognizing noise but it also misclassifies many data points as noise. DPC and K-means fail to recognize the clusters with noise.

As shown in \myfigref{Dartboard1}, the Dartboard1 dataset has four concentric rings. Both DPC-PPNNN and DBSCAN have perfect performance. DPC cannot find the correct cluster centers because the densities of the data points have little difference. Moreover, K-means has poor performance.

For the Jain dataset shown in \myfigref{Jain}, only DPC-PPNNN correctly identifies all clusters. DPC has not found the correct cluster centers because the difference between the densities of the two clusters is too large. The reason is the same for the poor performance of DBSCAN. K-means is not suitable for the datasets with streamline shapes.

\myfigref{R15} displays the results on the R15 dataset. The distribution of points makes it the most straightforward dataset for all the algorithms. Although there are some small defects among them, all the algorithms can recognize both the clusters and centers.

The clustering shown in \myfigref{Shapes} demonstrates the ability to address the datasets with clusters of different shapes. DPC-PPNNN, K-means and DBSCAN all perform well. Only DPC cannot detect the clusters because there is a cluster with manifold structure.

\subsection{Read-world datasets}
In this section, 8 UCI datasets, as shown in \mytabref{real-world}, are used to demonstrate the performance of the DPC-PPNNN clustering algorithm. These datasets are different in terms of sample number, feature number and cluster number. As shown in \mytabref{real-world-results}, DPC-PPNNN performs almost the best among the test cases.

\section{Conclusions} \label{conclusions-section}
In this paper, we proposed an improved probability propagation algorithm for density peak clustering based on natural nearest neighbors (DPC-NNN). The new algorithm does not require any parameters and can recognize cluster centers automatically. The final clustering process of the DPC-PPNNN motivated by the epidemic spread performs well especially for distinguishing two clusters that are close to each other. 

However, since the algorithm is based on density and probability, its performance is not good and stable enough, especially when all the data points have similar $\delta$ or similar densities $\rho$. For future work, we will work on solving these problems.

\section*{Acknowledgments}
The work was supported by National Natural Science Foundation of China (No. 12071453), the National Key R and D Program of China (2020YFA0713100),  Anhui Initiative in Quantum Information Technologies (AHY150200), and the Innovation Program for Quantum Science and Technology (2021ZD0302904).

% 参考文献
%\bibliographystyle{elsarticle-num} % 设置参考文献类型
%\bibliography{ref} % 生成参考文列表

\end{document}